\documentclass{article}
\usepackage{spconf,amsmath,graphicx}

\usepackage{epsfig}
\usepackage{graphicx}
\usepackage{amsmath}
\usepackage{amssymb}
\usepackage{times}
\usepackage{epsfig}
\usepackage{graphicx}
\usepackage{amsmath}
\usepackage{amssymb}
\usepackage{tabularx}
\usepackage{multirow}
\usepackage{subfigure}
\usepackage{comment}

\usepackage{xspace}
\makeatletter
\DeclareRobustCommand\onedot{\futurelet\@let@token\@onedot}
\def\@onedot{\ifx\@let@token.\else.\null\fi\xspace}
 
\def\ie{i.e\onedot}

\usepackage{booktabs}
\usepackage[pagebackref=true,breaklinks=true,letterpaper=true,colorlinks,bookmarks=false]{hyperref}

\title{FEATURELESS: BYPASSING FEATURE EXTRACTION IN ACTION CATEGORIZATION}
\twoauthors
  {S. L. Pintea$^{a}$, P. S. Mettes$^{a}$}
	{$^{a}$Intelligent Sensory Information Systems, \\
	University of Amsterdam, \\
 	Amsterdam, Netherlands}
  {J. C. van Gemert$^{a, b}$, A. W. M. Smeulders$^{a}$}
	{$^{b}$Computer Vision Lab, \\
         Delft University of Technology, \\
         Delft, Netherlands}

\begin{document}
\maketitle
\newcommand{\todo}[1]{\textcolor{red}{#1}}
\newcommand{\maybe}[1]{\textcolor{black}{#1}}
\begin{abstract}
This method introduces an efficient manner of learning action categories without the need of feature estimation.
The approach starts from low-level values, in a similar style to the successful CNN methods. 
However, rather than extracting general image features, we learn to predict specific video representations from raw video data.
The benefit of such an approach is that at the same computational expense it can predict 2$D$ video representations as well as 3$D$ ones, based on motion.
The proposed model relies on discriminative Waldboost, which we enhance to a multiclass formulation for the purpose of learning video representations.
The suitability of the proposed approach as well as its time efficiency are tested on the UCF11 action recognition dataset.
\end{abstract}
\begin{keywords}
Multiclass Waldboost, video representations, action recognition, feature learning.
\end{keywords}
\section{Introduction}
\label{sec:intro}
\begin{figure}
	\centering
	\includegraphics[width=0.96\linewidth]{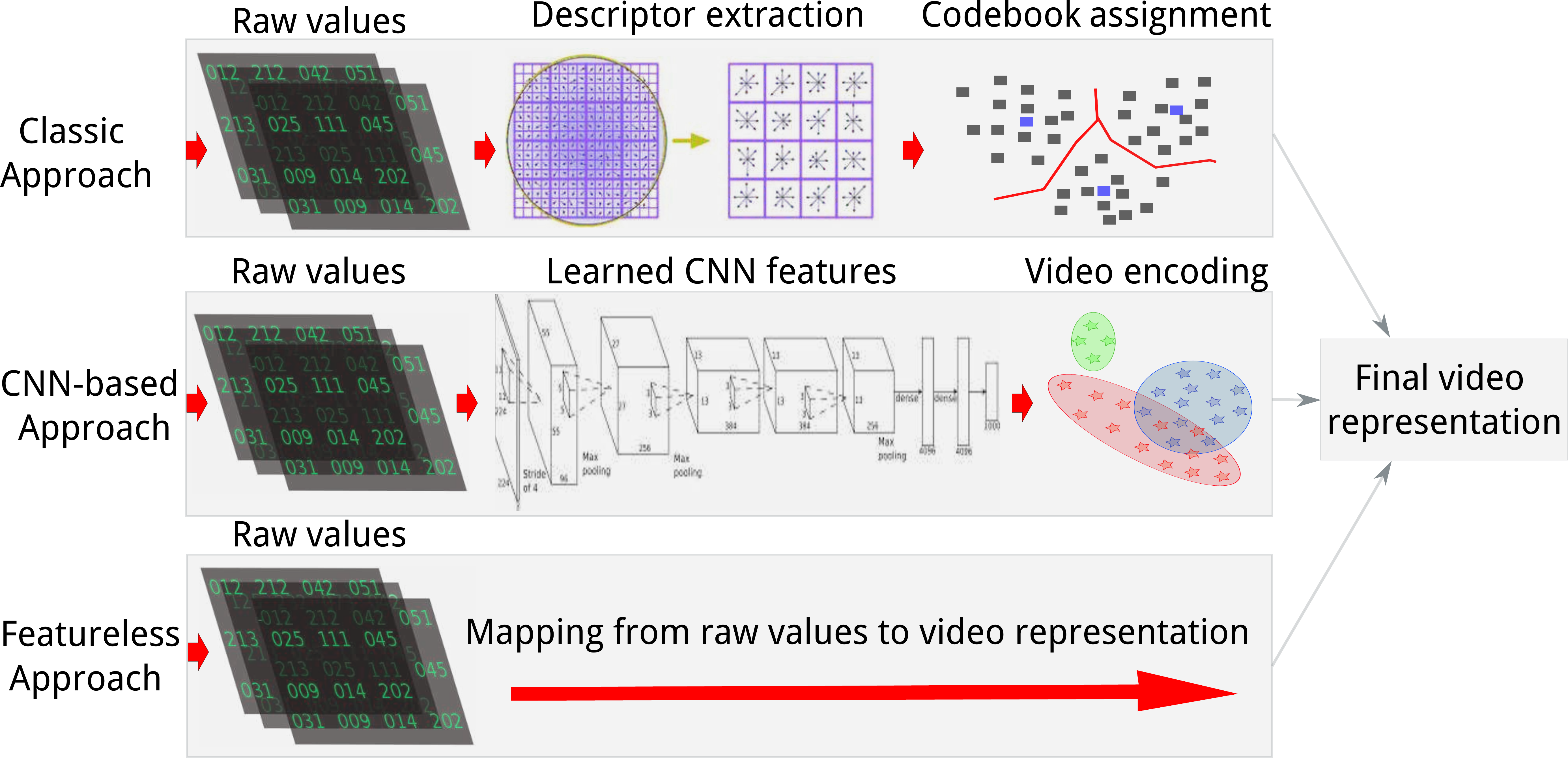}
	\caption{\small Going from raw video values to higher level features. 
		We propose bypassing feature computation by learning the mapping from raw input data to any type of higher order video representation --- 
		\ie based on appearance, motion descriptors.}
	\vspace{-15px}
	\label{fig:featureless}
\end{figure}

\vspace{-5px}
Ever since the bag-of-words representation of visual information was proposed \cite{sivic2003video}, the focus has been on feature robustness \cite{dalal2005histograms,lowe2004distinctive,van2010evaluating}.
The popular deep CNN (Convolutional Neural Networks) \cite{krizhevsky2012imagenet,schmidhuber2015deep,vondrick2015visualizing} have effectively replaced the handcrafted descriptors with network features.
Such networks have been successfully applied in the domain of action recognition \cite{karpathy2014large,cheron2015p,weinzaepfel2015learning,simonyan2014two}.
More recently, CNN features are used together with Fisher Vectors to build stronger video representation \cite{jain201515,xu2014discriminative}.
However, competitive performance for action recognition is still achieved by video representations relying on appearance and motion descriptors \cite{Wang2013,peng2014action,fernando2015modeling}.
This work proposes a manner of learning a given video representation, rather than learning better features to be subsequently used in the video representation, as in the case of CNN features.
Illustrated in figure~\ref{fig:featureless} is the proposed method that bypasses feature computation and, instead, learns the final video representation.

This work presents a proof of concept which challenges the idea of discarding the feature estimation and learning in one step the transition from low-level data to the final video representation.
The premise of this paper is to keep the advanced analysis as simple as possible, therefore, here we focus on the more simplistic bag-of-words model.
We research how much of this classic pipeline can be discarded while still achieving comparable performance. 
In the proposed method we learn from low-level values to predict a known codebook assignment --- thus, at test time neither the image descriptors, nor the codebook need to be defined.
Motion features as well as appearance features together with their codebook assignments are bypassed in the featureless method.

\begin{figure*}
	\centering
	\includegraphics[width=0.92\linewidth]{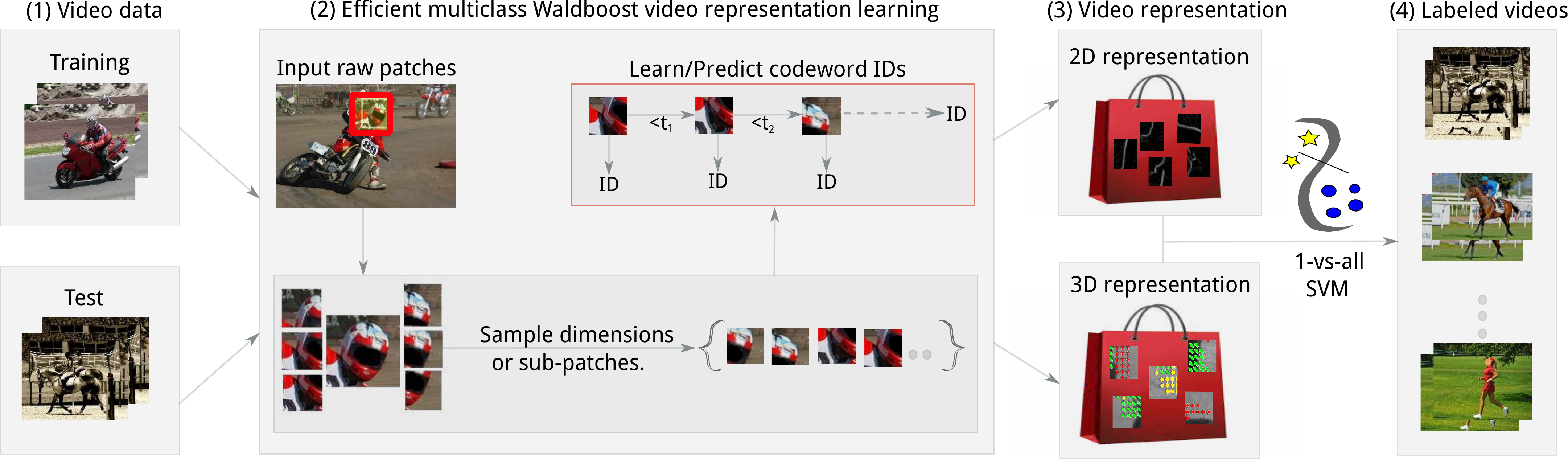}
	\caption{\small
		We start from input videos, we subsequently extract frame patches to be used as training samples for the multiclass Waldboost.
		During training we estimate appearance and motion descriptors --- HOG and HOF and 3$D$ HOF --- and we build a codebook which defines 
		the training labels.
		For each low-level patch extracted from an input video frame, we learn its final codebook assignment in the proposed multiclass Waldboost 
		extension.
		At test time we discard the codebook and predict in one step, from low-level patches, the final codebook assignment which is subsequently 
		input to an SVM for the final video categorization.}   
	\vspace{-15px}
	\label{fig:pipeline}
\end{figure*}
We rely on boosting to solve the learning problem.
The gain of starting from boosting techniques is their well known quality of being specifically appropriate for real time applications \cite{hall2014online,hall2014categories}.
This is a desirable property in the context of videos.
We start from individual values and build a strong classifier by incorporating multiple local cues in a boosting framework.
In this work, we propose a straightforward discriminative multiclass extension of Waldboost \cite{sochman2005waldboost}.
This is employed to learn the mapping from the low-level input visual information to the desired representation.

Tasks such as large scale video categorization and event recognition still rely on the use of large codebooks and descriptor extraction \cite{sundiscover,soomro2012ucf101}.
Moreover, the action recognition field deals with large amounts of data whose processing comes at considerable computational costs.    
This work can prove useful for such approaches, as it discards descriptor extraction and the need of codebooks or other video encodings such as Fisher Vectors at test time.
Figure~\ref{fig:pipeline} displays the proposed framework in the context of action recognition.
\section{Related Work}
\vspace{-5px}
In the literature the focus has been on either more compact and robust descriptors \cite{dalal2005histograms,lowe2004distinctive,van2010evaluating},
or on stronger video encodings \cite{jegou2012aggregating,sanchez2013image,van2010visual}.
The effective CNN methods are able to produce strong image features \cite{krizhevsky2012imagenet,schmidhuber2015deep,vondrick2015visualizing}.
Followed by a subsequent video encoding step, these methods represent the state-of-the-art in action recognition \cite{jain201515,xu2014discriminative}.
In this work we do not focus on learning either the image\slash frame features or the video encoding, but instead, we focus on learning a time-efficient 
classifier that bypasses these steps and retrieves the final video representation. 

We use as starting point the real-valued multiclass Adaboost \cite{zhu2009multi}. 
This is extended in a discriminative manner to determine at each iteration whether to continue with the evaluation of the next weak classifier or stop. 
Thus, it allows for fast decision making as not all weak classifiers need to be employed. 
As described in section~\ref{sec:wald}, this corresponds to a discriminative multiclass extension of Waldboost \cite{sochman2005waldboost}.  

In the action recognition literature, efficiency has been a main focus \cite{kantorovefficient,oneata:hal-00979594}.
Proposed methods are ranging from faster features, based on faster flow computation, to faster video encodings, based on additive approximations.
Methods such as \cite{cheron2015p,simonyan2014two,gkioxari2015contextual,sun2015human} have focused on improving the CNN architecture to achieve better 
performance in the context of action recognition.
Yet, with the gain in performance comes also a gain in speed as deep neural networks are known to be fairly efficient at test-time.
However, when focusing on the performance gain, methods \cite{Wang2013,peng2014action,fernando2015modeling} successfully rely on handcrafted visual descriptors such as HOG, HOF and MBH, and video encodings such as Fisher Vectors and VLAD.   
In this work we propose to combine the best of both worlds --- allow to obtain video representations similar to the ones based on visual descriptors such as HOF, HOG and MBH while not having to estimate these descriptors or define the video encoding.
Therefore, we bypass feature computation and video encoding and learn a direct mapping from low-level data.
\section{Learning the Mapping}
\vspace{-5px}
\subsection{Multiclass Extension of Walboost} 
\label{sec:wald}
\vspace{-5px}
\textbf{Data and Labels.} 
We use grayscale pixel values as input to the boosting pipeline. 
Given and input image patch, we perform $L_1$ normalization over the patch. 
During the boosting training we sample randomly $\sqrt{D}$ dimensions, where $D$ is the total number of data dimensions. 
For stability, we repeat the selection step $\sqrt{D}$ number of times and keep the feature dimensions that provide the best performance on the training data.
We use these dimensions to train a weak classifier. 
During training, each patch of gray values is associated with a descriptor which is subsequently projected on a codebook in order to retrieve the codebook assignment. 
The codebook assignment defines the labeling for the multiclass Waldboost.\\[5px]
\textbf{Weak Learners.} 
The weak learners are a set of $M$ multiclass decision trees with probabilistic outputs, $\mathbf{p}^m(\cdot)$, where $m \in \{1,.. M\}$ is the tree index. 
The maximum depth of each tree is set to 15, as standardly done in literature. 
Each decision tree predicts a $K$ dimensional probabilistic output, where $K$ is the number of classes --- codebook size. 
As suggested in~\cite{zhu2009multi}, we weight the decision boundary in each leaf by the current weights of the training samples falling in that leaf:
\begin{alignat}{1}
	{p}^m_k(\mathbf{x}_i) &= \frac{\sum_{i \in \mathcal{L}} w_i (y_i^k=1)}{\sum_{i \in \mathcal{L}} w_i}, \forall k \in \{1,.. K\},
\label{eq:weak}
\end{alignat}
where $\mathcal{L}$ --- the leaf reached by sample $\mathbf{x}_i$, $w_i$ --- the weight associated with $\mathbf{x}_i$, and $y_i^k$ the label of $\mathbf{x}_i$ and class $k$.
The weights of the training samples are only used in this step and reset for each weak learner.\\[5px] 
\textbf{Data Pool Sampling.} 
We follow the standard approach and initialize the weights, $\mathbf{w}$, with $\frac{1}{N}$, where $N$ is the number of training samples for the current 
weak classifier.
After training each weak classifier the weights are updated as follows:
\begin{alignat}{1}
	\label{eq:weights}
	w_i &= w_i \exp \left(-\frac{K-1}{K}\sum_k^K y_i^k \log p^m_k(\mathbf{x}_i) \right),
\end{alignat}
where the labels for the current sample are:
\begin{alignat}{1}
	y_i^{k^\prime} &= 
	\begin{cases}
	    1, & \text{if }k^\prime = k,\\
	    \frac{-1}{K-1}, & \text{otherwise}.
	\end{cases}
\end{alignat}
During training we sample a 10$^{th}$ of the complete number of training patches to be used for learning each weak classifier. 
Following \cite{kalal2008weighted}, we use the QWS+trimming (Quasi-random Weighted Sampling with trimming) for this step. 
Although each weak classifier is trained on a subset of the training data, the weight update (eq.~\ref{eq:weights}) is applied over the predictions of 
the weak classifier on the complete data.\\[5px]         
\textbf{Strong Classifier.} 
Following~\cite{zhu2009multi}, the probabilities of each weak classifier are transformed into real-valued scores:    
\begin{equation}
	\small
	s_k^m(\mathbf{x}_i) = (K-1) \left(\log p^m_k(\mathbf{x}_i) - \frac{1}{K} \sum_{k^\prime}^K 
		\log p^m_{k^\prime}(\mathbf{x}_i) \right).\\
	\label{eq:scores}
\end{equation} 
The final probabilistic prediction is obtained by taking the softmax over the sum of the scores of the weak classifiers:
\begin{alignat}{1}
	\label{eq:final_scores}
	s_k(\mathbf{x}_i) &= \sum_m^M s^m_k(\mathbf{x}_i),\\
	p_k(\mathbf{x}_i) &= \frac{\exp(s_k(\mathbf{x}_i))}{\sum_{k^\prime}^K \exp(s_{k^\prime}(\mathbf{x}_i))}.
\end{alignat} 
\textbf{Discrimiantive Multiclass Waldboost.} 
Rather than testing all the weak classifiers to reach a decision, we can determine if the strong classifier is sufficiently confident in its prediction 
and stop without evaluating the subsequent weak classifiers. 
Waldboost~\cite{sochman2005waldboost} selects a stopping threshold for each weak classifier.
These thresholds are learned over the strong classifier scores up to the current iteration. 

We propose an intuitive deterministic approach to learning stopping thresholds. 
After training each weak classifier on its current data pool, we employ it together with all the preceding weak classifiers to obtain a strong prediction on a validation set, denoted by $\bar{\mathbf{x}}$. 
The strong classifier scores (eq.~\ref{eq:final_scores}) up to the current step on the validation set are input to a new decision tree. 
We train one such stopping decision tree per weak classifier, $m$. 
The stopping classifiers return $K$ class probabilities.
We choose decision trees as stopping classifiers for both consistency as well as efficiency. 

At test time, the strong classifier scores up to the current step, $m$, are passed on to the current decision tree used for stopping. 
The stopping classifier decides whether to stop or continue with the estimation of the next boosting weak classifier. 
This is done by evaluating the output of the stopping classifier against a fixed desired class probability, $\alpha$:
\begin{alignat}{1}
	\label{eq:stop}
	\small
	\max_{k} \left(\text{Stop}_k^{M^\prime}\left(\sum_{m=1}^{M^\prime\le M} s^m_k(\bar{\mathbf{x}}_i)\right) \right) &\ge \alpha,
\end{alignat}    
where {\small $\text{Stop}_k^{M^\prime}(\cdot)$} --- stopping probability for class $k$ at $M^\prime$. 
\begin{table}
	\centering
	\fontsize{7.5}{8}\selectfont
	\begin{tabular}{ccc ccc}\toprule
			 & HOF    & Adaboost      & Waldboost     & 3$D$ HOF & Waldboost\\ 
			 &        & HOG\slash HOF & HOG\slash HOF &	      & 3$D$ HOF\\ \cmidrule(r){2-6}
	 Time\slash Frame       & 0.60 s & 4.00 s        & 0.60 s        & 7.50 s   & 0.60 s\\ \bottomrule
	\end{tabular}
	\caption{\small Run-times of the proposed featureless method versus descriptor extraction.
		The proposed method is on par with optimized HOF descriptor estimation and ten fold faster than 3$D$ HOF.}
	\vspace{-15px}
	\label{tab:times}
\end{table}
\subsection{Computational Requirements} 
\vspace{-5px}
The speed of dense descriptor extraction has improved considerably as most of the available implementations rely on integral images \cite{vedaldi08vlfeat}. 
However, when the temporal dimension is used --- 3$D$ descriptors, or motion information --- flow-based descriptors, the computational requirements increase.
Table~\ref{tab:times} shows the times for the computation of different descriptors as well as the Adaboost\slash Waldboost run-times for predicting on the patches of a complete frame. 
The runtime numbers are obtained running a single-thread, unoptimized implementation in C++.
The Waldboost prediction is as fast as the HOF descriptor extraction over the same frame. 
However, when predicting a codebook assignment based on 3$D$ motion features, boosting approaches are faster while Waldboost is one order of magnitude faster.
\section{Experiments}
\vspace{-5px}
We compare against the representations we learn from --- BOW (bag-of-words) with k-means codebooks. 
The codebooks are extracted over HOF, HOG and 3$D$ HOF descriptors and used to define the multiclass Waldboost labels.
More powerful representation such as Fisher encodings or just the first fully connected layer in a pretrained CNN could also be used, 
by applying a discretization step. 
However, the aim is to keep the analysis simple and test the feasibility of transitioning in one step from the low-level features to the final video representation. 
Therefore, in our experimental setup we use small codebooks and input grayscale patches.

\begin{table*}[t!]
	\centering
	\small
	\begin{tabular}{llll lllp{.20\textwidth}l}\toprule
		        & \multicolumn{2}{c}{HOG} & \multicolumn{2}{c}{HOF} & \multicolumn{3}{c}{3$D$ HOF}\\ \cmidrule(r){2-3}\cmidrule(r){4-5}\cmidrule(r){6-8}
			    & BOW           & Waldboost & BOW           & Waldboost & BOW  & Waldboost & BOW \& Waldboost\\ \cmidrule(r){2-8} 
	MAP        & \textbf{44}\% & 41\%      & \textbf{37}\% & 32\%      & 45\% & 36\%      & \textbf{50}\% \\ \bottomrule\\
	\end{tabular}
	\caption{\small MAP (mean average precision) scores on UCF11 when using the standard BOW video representation as compared to the representation obtained from the 
		proposed multiclass Waldboost predictions. 
		The results of the featureless approach are comparable with the baseline, while the combination of the two gains in performance.}
	\vspace{-15px}
	\label{tab:featureless}
\end{table*}
\subsection{Experimental Setup} 
\vspace{-5px}
\textbf{Experiment 1} employs codebooks of only 100 dimensions obtained by applying k-means over 100,000 descriptors and compare 
against the standard BOW baseline.
For the 3$D$ descriptors we use larger codebooks --- 1000 $D$ --- which are also obtained by employing k-means over a set of 100,000 training descriptors.
The 3$D$ motion descriptors are computed over 8 pairs of frames.
\textbf{Experiment 2} discards the codebook altogether from both training and test and consider each descriptor to be its own cluster center. 
All experiments use 1000 weak classifiers, each trained on 24 randomly sampled data dimensions.
The patch sizes for descriptor extraction as well as for boosting are of 24$\times$24 px. 
For the multiclass Waldboost we set the stopping threshold, $\alpha$, to $.97$ as this proved effective in practice. 
All experiments report MAP scores (Mean Average Precision) on UCF11 \cite{liu2009learning} action recognition dataset.
\subsection{Experiment 1: Featureless}
\vspace{-5px}
\subsubsection{Experiment 1.1: Waldboost vs. Other Algorithms}
\label{subsec:exp11}
\begin{table}
	\centering
	\small
	\begin{tabular}{l*{4}{p{.10\textwidth}}}\toprule
		         & Linear SVM & Adaboost & Waldboost\\ \cmidrule(r){2-4}
	MAP              & 16\% & 41\% & \textbf{41}\% \\ \midrule
	Time\slash frame & 15.00 sec & 4.00 sec & 0.60 sec\\ \bottomrule
	\end{tabular}
	\caption{\small Performance of different learning algorithms when learning the mapping from input grayscale values to the HOG codebook assignment. 
		The proposed multiclass  Waldboost method achieves better performance than linear SVM and at the same time gains in efficiency over 
		Adaboost at no loss in performance.}
	\vspace{-15px}
	\label{tab:learn}
\end{table}
Table~\ref{tab:learn} displays comparative results when learning the mapping from grayscale input values to the HOG codebook assignment. 
Waldboost manages to outperform by a large margin the linear SVM classifier as it focuses the learning on the informative data dimensions.
At the same time, the proposed multiclass Waldboost brings gain in efficiency at no loss in performance when compared with Adaboost, although it analyzed only a subset of the weak classifiers.
\subsubsection{Experiment 1.2: Learning vs. Feature Extraction}
\label{subsec:exp12}
This experiment tests the feasibility of the featureless aim. 
Given a set of grayscale input patches together with their codebook assignments over the associated descriptors, it trains a multiclass Waldboost. 
At test time no descriptors or codebooks are used, thus, obtaining a featureless representation. 
Table~\ref{tab:featureless} depicts the results obtained by the proposed method when compared to the classic BOW approach. 
The performance of BOW is slightly better than Walboost. 
The work of \cite{reddy2013recognizing} reports an accuracy of 55.46\% on BOW with SVM and SIFT descriptors on a codebook of 500 dimensions. 
Our methods based on BOW and Walboost over HOG features using a codebook of only 100 dimensions obtains a competitive accuracy of 43.18\% which corresponds to a 
mean average precision of 41\%, as listed in table~\ref{tab:featureless}. 
However, when combining the two representation --- as in the case of 3$D$ HOF, the combined representation exceeds in performance both BOW and Waldboost.
This indicates that despite using the same starting point --- the same codebook and descriptor assignment, the BOW and Waldboost representations encode complementary information.   
\subsection{Experiment 2: Featureless and Codebookless}
\label{subsec:exp3}
\vspace{-5px}
\begin{table}
	\centering
	\small
	\begin{tabular}{llll}\toprule
		   & BOW        & \multicolumn{2}{c}{Codebookless}\\ \cmidrule(r){3-4}
		   &            & Adaboost & Waldboost\\ \cmidrule(r){2-4}
	MAP        & 44\%       & 41\%     & 37\% \\ \bottomrule
	\end{tabular}
	\caption{\small MAP scores on UCF11 for the BOW baseline on a 100$D$ codebook as well codebookless Adaboost and Waldboost --- no codebooks 
		are used during training. 
		The learned mapping predicts patch IDs rather than codebook IDs. 
		The boosting methods manage to achieve comparable performance to the baseline despite discarding both descriptors and codebooks.}
	\vspace{-15px}
	\label{tab:codebookless}
\end{table}
Table~\ref{tab:codebookless} displays the action recognition performance when the boosting techniques learn from both a featureless as well as 
codebookless representation. 
Each patch is considered to be the center of its own cluster, thus the mapping learns to predict patch IDs rahter than codebook IDs.
Out of the 100,000 patches considered, only $\approx$100 unique patch IDs are begin predicted, the rest having zero predictions.
Both Adaboost as well as the proposed multiclass Waldboost still manage to learn the underlying structure in the data, despite not making use 
of either descriptors or codebooks at test-time. 
\section{Conclusions}
\vspace{-5px}
This work analyzes whether we can bypass feature extraction and still attain comparable performance with the framework we learn from.
In search for the simplest possible method, we learn a mapping from grayscale values to existing representations such as codebooks. 
A straightforward multiclass extension of Waldboost is brought forth for learning this mapping. 
The efficiency as well as the performance of the proposed method are tested in the context of action recognition.
Moreover, we also consider video representations based on motion features, as well as discarding the codebook altogether and learning 
both a featureless and codebookless mapping.\\[5px]  
\small
\textbf{Acknowledgments}
This research is supported by the Dutch national program COMMIT.
\bibliographystyle{IEEEbib}
\bibliography{featureless}

\begin{thebibliography}{10}

\bibitem{sivic2003video}
J.~Sivic and A.~Zisserman,
\newblock ``Video google: A text retrieval approach to object matching in
  videos,''
\newblock in {\em ICCV}, 2003.

\bibitem{dalal2005histograms}
N.~Dalal and B.~Triggs,
\newblock ``Histograms of oriented gradients for human detection,''
\newblock in {\em CVPR}, 2005.

\bibitem{lowe2004distinctive}
D.~Lowe,
\newblock ``Distinctive image features from scale-invariant keypoints,''
\newblock {\em IJCV}, 2004.

\bibitem{van2010evaluating}
K.~Van De~Sande, T.~Gevers, and C.~Snoek,
\newblock ``Evaluating color descriptors for object and scene recognition,''
\newblock {\em PAMI}, 2010.

\bibitem{krizhevsky2012imagenet}
A.~Krizhevsky, I.~Sutskever, and G.~E. Hinton,
\newblock ``Imagenet classification with deep convolutional neural networks,''
\newblock in {\em NIPS}, 2012.

\bibitem{schmidhuber2015deep}
J.~Schmidhuber,
\newblock ``Deep learning in neural networks: An overview,''
\newblock {\em Neural Networks}, 2015.

\bibitem{vondrick2015visualizing}
C.~Vondrick, A.~Khosla, H.~Pirsiavash, T.~Malisiewicz, and A.~Torralba,
\newblock ``Visualizing object detection features,''
\newblock {\em CoRR}, 2015.

\bibitem{karpathy2014large}
A.~Karpathy, G.~Toderici, S.~Shetty, T.~Leung, R.~Sukthankar, and L.~Fei-Fei,
\newblock ``Large-scale video classification with convolutional neural
  networks,''
\newblock in {\em CVPR}, 2014.

\bibitem{cheron2015p}
G.~Ch{\'e}ron, I.~Laptev, and C.~Schmid,
\newblock ``P-cnn: Pose-based cnn features for action recognition,''
\newblock in {\em ICCV}, 2015.

\bibitem{weinzaepfel2015learning}
P.~Weinzaepfel, Z.~Harchaoui, and C.~Schmid,
\newblock ``Learning to track for spatio-temporal action localization,''
\newblock in {\em CVPR}, 2015.

\bibitem{simonyan2014two}
K.~Simonyan and A.~Zisserman,
\newblock ``Two-stream convolutional networks for action recognition in
  videos,''
\newblock in {\em NIPS}, 2014.

\bibitem{jain201515}
M.~Jain, J.~van Gemert, and C.~Snoek,
\newblock ``What do 15,000 object categories tell us about classifying and
  localizing actions?,''
\newblock in {\em CVPR}, 2015.

\bibitem{xu2014discriminative}
Z.~Xu, Y.~Yang, and A.~Hauptmann,
\newblock ``A discriminative cnn video representation for event detection,''
\newblock {\em CoRR}, 2014.

\bibitem{Wang2013}
H.~Wang and C.~Schmid,
\newblock ``Action recognition with improved trajectories,''
\newblock in {\em ICCV}, 2013.

\bibitem{peng2014action}
X.~Peng, C.~Zou, Y.~Qiao, and Q.~Peng,
\newblock ``Action recognition with stacked fisher vectors,''
\newblock in {\em ECCV}. 2014.

\bibitem{fernando2015modeling}
B.~Fernando, E.~Gavves, J.~Oramas, A.~Ghodrati, and T.~Tuytelaars,
\newblock ``Modeling video evolution for action recognition,''
\newblock in {\em CVPR}, 2015.

\bibitem{hall2014online}
D.~Hall and P.~Perona,
\newblock ``Online, real-time tracking using a category-to-individual
  detector,''
\newblock in {\em ECCV}. 2014.

\bibitem{hall2014categories}
D.~Hall and P.~Perona,
\newblock ``From categories to individuals in real time—a unified boosting
  approach,''
\newblock 2014.

\bibitem{sochman2005waldboost}
J~Sochman and J.~Matas,
\newblock ``Waldboost-learning for time constrained sequential detection,''
\newblock in {\em CVPR}, 2005.

\bibitem{sundiscover}
C.~Sun and R.~Nevatia,
\newblock ``Discover: Discovering important segments for classification of
  video events and recounting,''
\newblock in {\em CVPR}, 2014.

\bibitem{soomro2012ucf101}
K.~Soomro, A.~Zamir, and M.~Shah,
\newblock ``Ucf101: A dataset of 101 human actions classes from videos in the
  wild,''
\newblock {\em CoRR}, 2012.

\bibitem{jegou2012aggregating}
H.~J{\'e}gou, F.~Perronnin, M.~Douze, J.~S{\'a}nchez, P.~P{\'e}rez, and
  C.~Schmid,
\newblock ``Aggregating local image descriptors into compact codes,''
\newblock {\em PAMI}, 2012.

\bibitem{sanchez2013image}
J.~S{\'a}nchez, F.~Perronnin, T.~Mensink, and J.~Verbeek,
\newblock ``Image classification with the fisher vector: Theory and practice,''
\newblock {\em IJCV}, 2013.

\bibitem{van2010visual}
J.~van Gemert, C.~Veenman, A.~Smeulders, and J.~Geusebroek,
\newblock ``Visual word ambiguity,''
\newblock {\em PAMI}, 2010.

\bibitem{zhu2009multi}
J.~Zhu, H.~Zou, S.~Rosset, and T.~Hastie,
\newblock ``Multi-class adaboost,''
\newblock {\em Statistics and its Interface}, 2009.

\bibitem{kantorovefficient}
V.~Kantorov and I.~Laptev,
\newblock ``Efficient feature extraction, encoding and classification for
  action recognition,''
\newblock in {\em CVPR}, 2014.

\bibitem{oneata:hal-00979594}
D.~Oneata, J.~Verbeek, and C.~Schmid,
\newblock ``Efficient action localization with approximately normalized fisher
  vectors,''
\newblock in {\em CVPR}, 2014.

\bibitem{gkioxari2015contextual}
G.~Gkioxari, R.~Girshick, and J.~Malik,
\newblock ``Contextual action recognition with r* cnn,''
\newblock {\em CoRR}, 2015.

\bibitem{sun2015human}
L.~Sun, K.~Jia, D.~Yeung, and B.~Shi,
\newblock ``Human action recognition using factorized spatio-temporal
  convolutional networks,''
\newblock in {\em ICCV}, 2015.

\bibitem{kalal2008weighted}
Z.~Kalal, J.~Matas, and K.~Mikolajczyk,
\newblock ``Weighted sampling for large-scale boosting,''
\newblock 2008.

\bibitem{vedaldi08vlfeat}
A.~Vedaldi and B.~Fulkerson,
\newblock ``Vlfeat: An open and portable library of computer vision
  algorithms,'' \url{http://www.vlfeat.org/}, 2008.

\bibitem{liu2009learning}
J.~Liu, Y.~Yang, and M.~Shah,
\newblock ``Learning semantic visual vocabularies using diffusion distance,''
\newblock in {\em CVPR}, 2009.

\bibitem{reddy2013recognizing}
K.~Reddy and M.~Shah,
\newblock ``Recognizing 50 human action categories of web videos,''
\newblock {\em Machine Vision and Applications}, 2013.

\end{thebibliography}
\end{document}